# UN-AVOIDS: Unsupervised and Nonparametric Approach for Visualizing Outliers and Invariant Detection Scoring

Waleed A. Yousef, *Senior Member, IEEE*, Issa Traoré, *Senior Member, IEEE*, and William Briguglio, *Member, IEEE*

*Abstract*—The visualization and detection of anomalies (outliers) are of crucial importance to many fields, particularly cybersecurity. Several approaches have been proposed in these fields, yet to the best of our knowledge, none of them has fulfilled both objectives, simultaneously or cooperatively, in one coherent framework. Moreover, the visualization methods of these approaches were introduced for explaining the output of a detection algorithm, not for data exploration that facilitates a standalone visual detection. This is our point of departure in introducing UN-AVOIDS, an unsupervised and nonparametric approach for both visualization (a human process) and detection (an algorithmic process) of outliers, that assigns invariant anomalous scores (normalized to $[0, 1]$), rather than hard binary-decision. The main aspect of novelty of UN-AVOIDS is that it transforms data into a new space, which is introduced in this paper as neighborhood cumulative density function (NCDF), in which both visualization and detection are carried out. In this space, outliers are remarkably visually distinguishable, and therefore the anomaly scores assigned by the detection algorithm achieved a high area under the ROC curve (AUC). We assessed UN-AVOIDS on both simulated and two recently published cybersecurity datasets, and compared it to three of the most successful anomaly detection methods: LOF, IF, and FABOD. In terms of AUC, UN-AVOIDS was almost an overall winner with a margin that varied between -0.028 and 0.125, depending on the data. The article concludes by providing a preview of new theoretical and practical avenues for UN-AVOIDS. Among them is designing a visualization aided anomaly detection (VAAD), a type of software that aids analysts by providing UN-AVOIDS' detection algorithm (running in a back engine), NCDF visualization space (rendered to plots), along other conventional methods of visualization in the original feature space, all of which are linked in one interactive environment.

## I. Introduction

The last decade has witnessed an exponential increase in the velocity, volume, and diversity of security data. Although this creates significant opportunity for improved threat hunting, it raises considerable challenges for security analysts, who are confronted with a large number of indicators of compromise (IOC), too many false positives (FPs), and a wide variety of security logs and data feeds. Faced with the aforementioned challenges, both detection algorithms and visualization methods represent important assets in achieving better monitoring and allowing security analysts to gain better insights into network traffic, master the complexity of the threat intelligence feeds, and achieve timely remediation of security incidents. Although the motivation of this article stems from cybersecurity, its utility and applicability are ubiquitous wherever data is acquired and anomaly (outlier) detection is crucial.

Several approaches (reviewed in Section II) have been proposed in different studies for either anomaly detection or anomaly visualization. To the best of our knowledge, each of these approaches is either a detection algorithm or a visualization method; in addition, the latter is designed for explanation, not for exploration. Said differently, these visualization methods rely on an underlying anomaly detection algorithm that performs the anomaly detection usually in the original feature space, which is almost always high dimensional. Next, the information on the detected anomalies is visualized, either in the original feature space or in other higher-level contexts. This visualization conveys to the analysts—who are not acquainted with either the mechanism or working space of the detection algorithm—enough of an explanation to interpret these algorithms' decisions.

We propose a novel approach for anomaly detection and visualization by combining both detection algorithms and exploratory visualization, where both can work independently or jointly in hunting for threats, finding anomalies in security data, or verifying whether alerts are FPs or true positives (TPs); in the end, this can help increase the number of (TPs) by quickly exposing novel or unexpected intrusion patterns. Both visualization (a human process) and detection (an algorithmic process) are conducted in the same space, which is later referred to as the neighborhood cumulative distribution function (NCDF) space, a two-dimensional (2D) space. This space is transformed from the original feature space without dimension reduction; rather, it is equivalent to the probability space of the cumulative density function (CDF). In this space, which is first constructed from the data before any visualization or detection step, each observation is represented as a 2D curve, where the curves of the anomalous observations become distant and distinguishable from those of the normal observations. In this space, visualization takes place by the analyst simply by observing the anomalous curves that have wide gaps separating them from the other curves. On the other hand, detection takes place algorithmically by assigning a high score to these anomalous curves. Said differently, the visual exploration pursued by the analyst and the scoring calculations conducted by the detection algorithm are, respectively, the qualitative and quantitative measures for identifying the outliers. This key feature of UN-AVOIDS is very important for the human observer (the analyst), who uses the visualization process because it provides a clear intuition on how the algorithm works and makes its decisions. In this manner, both detection and visualization can be carried out jointly, i.e., cooperatively, or independently from each other.

Our approach was assessed on both simulated and real datasets. In addition to its unique visualization characteristics, its detection performance on the real datasets was compared with three cutting-edge anomaly detection methods, and it was an overall winner. Our proposed approach is an unsupervised and nonparametric approach for visualizing outliers by invariant detection scoring (UN-AVOIDS) that is:

**unsupervised**, where testing is pursued directly with no prior training. It does not assume any data labeling for any subset of the data, which implies that even single-class labeling is not assumed, as opposed to some other algorithms.

**nonparametric**, in two senses: (1) It does not assume, or take as an input, a probability distribution of the data. (2) It is parameter free, where there is no tuning or adjusting parameter that controls the accuracy of the algorithm.

**an approach for visualizing outliers**, where the analyst can explore the dataset and visually detect the outliers. Moreover, the visualization can be provided through any data visualization software as an interactive plot that is linked to other interactive plots. Such software enables the analyst to perform elaborate data visualization exploration and manipulations.

**an invariant detection scoring**, where the detection algorithm provides a score that is normalized to $[0, 1]$, which expresses the level of anomalousness (or outlierness) of an observation, as opposed to a hard-threshold decision.

The remainder of the paper is structured as follows. In Section II, we summarize and discuss the related work on both algorithmic and visualization approaches for anomaly detection and visualization, respectively. In Section III, we provide some important mathematical background that paves the way for our anomaly detection approach, and present UN-AVOIDS' mathematical framework and algorithm. In Section IV, we explain the experimental assessment of UN-AVOIDS on simulated and real datasets, present the results with many illustrations, and interpret these results. In Section V, we provide an extensive discussion and remarks on UN-AVOIDS, including computational power, complexity, and potential future extensions. Finally, Section VI concludes the article, and many mathematical properties and proofs are deferred to the Appendix. The majority of the figures in this paper include very dense information; therefore, they are produced as large-size, scalable vector graphics (SVG) that support infinite zooming.

## II. Related Work

### A. Literature Review of Algorithmic Approaches

Following the traditional paradigm of ML, outlier detection methods can be classified into two categories: supervised and unsupervised. However, for the supervised approach, some challenges may arise, including the class imbalance problem (because normal data are more abundant than anomalous data) and the partially observed class problem (where the training data may only be available from the normal class and not the anomalous one). On the other hand, many unsupervised methods are the one-class generalization to existing supervised methods. Before reviewing the literature, it is worth mentioning that although outlier detection and removal are performed prior to prediction, both can be performed simultaneously by designing the appropriate loss function. This approach can be pursued, e.g., in graph representation problems, by learning the deep neural network (DNN) parameters [1], or by solving the optimization problem to find the best embedding [2].

A comprehensive survey on anomaly detection algorithms can be found in [3–5]; in particular, a survey on using deep learning techniques is found in [6]; and surveys on anomaly detection for only categorical variables are found in [7–11]. Furthermore, there are surveys on anomaly detection in particular fields and applications, such as astrophysics [12], environmental data analysis [13], IoT [14, 15], network anomaly detection [16–21], and data mining [22]. In addition to these comprehensive review articles, we review next some research that is of particular interest to the present article.

[23] introduced an interesting approach for detecting anomalies in network traffic. The approach is cited in several publications and leveraged by systems and applications in cybersecurity, e.g., [24–28]. However, this approach, along with these citing papers, seem to form an isolated cluster of research work that is neither cited by, nor citing, other clusters of outlier detection approaches. Even more surprising, [29] pursued a very similar route and reached a very similar approach without citing the former. The approach of [23], in a simplified description, provides a score for an observation to express its anomalousness. The score expresses the relative probability of this observation with respect to (w.r.t.) the probabilities of others, rather than its absolute probability. Hence, the authors called it probability of probability, so it is meaningful and comparable across different datasets. To calculate the probability of an observation, the authors fit a mixture of Gaussian probability density, which casts the approach as a parametric model.

There is a subclass of anomaly detection algorithms that follow proximity-based approaches. These approaches investigate the $K$ nearest neighbors in some way or another, including cluster discovery, density estimation, distance calculation, and so forth, to test the closure of the point of interest (POI) to others, hence deciding on its anomalousness. A very well-known method in this category is the local outlier factor (LOF) [30], which is implemented in the Python library `scikit` [31]. Later, several variants of this approach were developed. Simply put, the main idea of this approach is that it compares the relative distance of the POI to the other distances of its local $K$ nearest neighbors. Based on this comparison, it assigns a LOF score to that observation to express its anomalousness.

[32] introduced isolation forest (IF) that creates an "isola-

tion tree" to recursively partition the dataset using a series of random cut values and randomly selected features. If the POI is isolated in its partition after a small number of partitions, i.e., near the root of the tree, it is likely an outlier because anomalies are more susceptible to isolation using random partitions.

[33] introduced the angle-based outlier detector (ABOD), which uses the variance of the angle, whose vertex is a particular POI, and whose sides connect to a pair of arbitrary points. Across all possible pairs, the variance of the angle tends to be very low for outliers. Because repeating this process for all possible pairs for all POIs can be computationally expensive, the authors also introduced fast-ABOD (FABOD), which only uses a subset of points to create the pairs.

[34] introduced `hdoutliers` to detect outliers in a dataset with mixed categorical and continuous features (after encoding categorical features using correspondence analysis), in very high dimensions up to hundreds of thousands (using random projection), and with a very large number of observations (using a leader algorithm for selecting exemplars). We classify `hdoutliers` as belonging to the proximity-based approach because it calculates the nearest neighbor distance pairwise. [35] tried to improve on `hdoutliers`, providing their elaboration `stray`, an open source R package.

### B. Literature Review of Visualization Approaches

As introduced in Sec. I, to the best of our knowledge, all existing methods for anomaly visualization have been designed for explanation, not for exploration. In addition, we are not aware of a comprehensive review of these visualization methods. [36], for example, reviewed interactive machine learning (IML) research, where some of the cited research relates to anomaly detection. Below, we review the literature that is most relevant to the present article.

[37] used parallel coordinates (∥-cords for short) to render a massive (up to three million observations) high-dimensional (up to 50 dimensions) dataset after emphasizing outliers (the reader may refer to [38], [39] for the full details on ∥-cords). The authors leveraged the power of ∥-cords to plot massive high-dimensional data to allow analysts to observe outliers in their original feature space. However, because the few outliers will most certainly be obscured by the massive number of observations, the authors first detected the outliers using a simple 2D histogram algorithm, which calculates the observations' density in a 2D binning mesh for each pair of ∥-cords. Then, they plotted the discovered outliers (those observations that contribute to the low-density bins) with more emphasis (color/brightness) on the ∥-cords.

[40] presented APT-Hunter, a visualization method that is very specific for detecting malicious logins in enterprise networks. It builds a node-link diagram to illustrate the users' connectivity based on the administrator's queries and search. From the topology of the diagram, and the three main credentials (user, source, and destination) drawn by the tool, the administrators can detect any anomalous behavior using their experience and background domain knowledge. This visualization method is a very high-level representation of the data, at the system level, rather than visualization of the data in its plain $p$-dimensional feature space.

[41] introduced their system, TargetVue, for visual analysis and detection of users' anomalous behavior in online communication systems. The underlying anomaly detection algorithm of TargetVue is LOF [30]. TargetVue selects the highest $n$ scores provided by LOF and visualizes the $p$-dimensional feature vectors of the corresponding $n$ users using their novel visualization method, Z-glyph, which will be detailed below. Similar to all visualization glyphs, each Z-glyph has a unique 2D shape that provides a one-to-one map with the original feature vector. When data analysts investigate the $n$ Z-glyphs, they can understand the features and characteristics that nominated them, by LOF, to be outliers. Having said that, TargetVue is not a detection algorithm in its own right; rather, it intuitively visualizes the anomalies detected by LOF so that analysts may interpret the reason behind them being anomalous.

After its introduction by [41], [42] elaborated on Z-glyph and presented it as a visualization and detection method for humans, who can detect the anomalies by looking at the whole $n$ Z-glyphs of a dataset of size $n$. From this perspective, this method is suitable for any $p$-dimensional dataset, not only for those acquired from the network communication domain. Although there are many types of visualization glyphs [43], Z-glyph has a special feature: it visualizes each observation after normalization with respect to the whole dataset to resemble the baseline behavior of communication networks. Therefore, the outlierness of an observation is more obvious for human observers. The main disadvantage of using glyphs for outlier detection, whether it is the Z-glyph or others, is that it is almost impossible for any human to look at $n > 100$, let alone millions of glyphs, to identify outliers.

[44] introduced EVA, a system for financial fraud detection (FFD) and visualization. The system has its own anomaly detection algorithm, whose details are not revealed by the authors. The algorithm provides an overall anomaly score for each observation (activity) and an individual score for each feature of this activity. Next, several kinds of interactive plots (via linking and brushing) are provided for activity visualization.

[45] introduced the platform Situ for the detection and visualization of suspicious network behavior. Although their application domains are different, Situ, EVA, and TargetVue are conceptually very similar in that the three of them call an underlying algorithm to detect anomalies in the dataset and then visualize them so that analysts can understand and interpret the reason behind this anomalousness. However, Situ is different in two ways. First, the underlying anomaly detection algorithm comes from [26]. Second, the visualization is both comprehensive and contextual, and uses many pallets and plots at the system level to aid administrators in making a decision.

### III. UN-AVOIDS Outliers

Our detection and visualization approach is based on mapping the feature space to a new space that we introduce: the NCDF, a name inspired by the cumulative distribution function (CDF) of a random variable (r.v.). Since an observation may be anomalous w.r.t. a local neighborhood and normal w.r.t. others, the NCDF space is a characteristic of all observations w.r.t. to all possible neighborhoods under a particular norm. The most important aspect of this new 2D space is that it is not a projection from $\mathbf{R}^d$ onto $\mathbf{R}^2$. Rather, it is a lossless transformation of the probability space, as explained next. We start with formal definitions and properties of the new NCDF space (Section III-A), which establish the cornerstone of the visualization and detection approach of UN-AVOIDS, which is explained next in Section III-B.

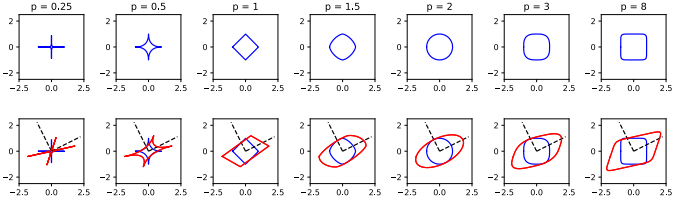

Figure 1: The neighborhood $\mathcal{N}_{\|\cdot\|_p, A}(x_0, \epsilon)$ for $x_0 = 0$, $\epsilon = 1$, and different values of $p$. Top: $A = I$; bottom: $A = \lambda_1 v_1 v_1' + \lambda_2 v_2 v_2'$, $v_1 = (2,1)'/\sqrt{5}$, $v_2 = (-1,2)'/\sqrt{5}$, $\lambda_1 = 2$, $\lambda_2 = 1$.

### A. Mathematical Framework: Definitions and Properties

**Definition 1.** *The closed neighborhood (or ellipsoid) of radius $\epsilon$, under the general norm $\|\cdot\|$ and a transformation matrix $A$, centered around $x_0 \in \mathbf{R}^d$ is defined as:*

$$\mathcal{N}_{\|\cdot\|, A}(x_0, \epsilon) = \{x \mid \|A^{-1}(x - x_0)\| \le \epsilon\}. \tag{1}$$

A special norm of interest is the $L^p$-norm, $\|x\|_p = (\sum_i |x_i|^p)^{1/p}$. Figure 1 is an illustration of $\mathcal{N}_{\|\cdot\|_p, A}(0, 1)$ in $\mathbf{R}^2$ and different values of $p$. The top and second rows correspond to $A = I$ (identity matrix), and $A = \lambda_1 v_1 v_1' + \lambda_2 v_2 v_2'$, $v_1 = (2,1)'/\sqrt{5}$, $v_2 = (-1,2)'/\sqrt{5}$, $\lambda_1 = 2$, $\lambda_2 = 1$, respectively. For these two values of $A$, and for the special value of $p = 2$, $\mathcal{N}$ is a circle or ellipsoid, respectively. It is clear that the orientation and axes' length of the general ellipsoid in $\mathbf{R}^d$ are determined by the eigenvectors $v_i$s and the eigenvalues $\lambda_i$s of $A$, respectively. To simplify the notation, the subscript $A$ will be dropped when $A = I$. The following remark establishes a foundation of UN-AVOIDS:

**Remark 1.**
1) *For $p < 1$, $\mathcal{N}$ is always non-convex (Lemma VII). In the vernacular, regardless of the dimensionality $d$ and data transformation $A$, the neighborhood with $p < 1$ will have spikes that become sharper with a smaller $p$.*
2) *In high dimensions (the curse of dimensionality), it is well known that almost all the data will be on the surface of a hypercube. In that case, when $1 \le p$, any neighborhood of a particular volume, inside the hypercube, will be almost empty. However, for $p << 1$, a neighborhood with spikes can expand, without consuming the Euclidean volume, and hunt data scattered on the hypercube surface.*

**Definition 2** ($NCDF_{\|\cdot\|_p, X=x}$: population NCDF). *Consider the r.v. $X \in \mathbf{R}^d$, its CDF $F_X$, and a point $x$. The NCDF for $X$ around $x$ under $\|\cdot\|_p$ is given by:*

$$NCDF_{\|\cdot\|_p, X=x}(v) = \Pr[\mathcal{N}_{\|\cdot\|_p}(x, V_{\|\cdot\|_p}^{-1}(v))] \tag{2a}$$

$$= \int_{\mathcal{N}} dF_X, \tag{2b}$$

*where $V_{\|\cdot\|_p(\epsilon)}$ is the volume of the neighborhood $\mathcal{N}_{\|\cdot\|_p}(x, \epsilon)$, and (2b) only exists if $F_X$ is differentiable (i.e., has a PDF).*

**Remark 2.**
1) *The NCDF of a r.v. $X$, around an observation $x$, is a function of the probability of a neighborhood under $\|\cdot\|_p$ vs. its volume $v = V_{\|\cdot\|_p(\epsilon)}$ (not vs. its radius $\epsilon = V^{-1}_{\|\cdot\|_p}(v)$).*
2) *This definition of the NCDF is not explicit to $\|\cdot\|_p$; it can be adopted for any other general norm $\|\cdot\|$. However, another norm $\|\cdot\|$ must be defined before being able to find a relationship between the neighborhood radius and volume (see Lemma 8).*
3) *We defined the NCDF to be the probability vs. the volume $v$ of a neighborhood, rather than its radius $\epsilon$, for two reasons. First, this can provide a general definition that is applicable to other forms of norms that may imply several parameters to describe a neighborhood rather than a simple radius. Second, although there is a one-to-one mapping between the volume of the neighborhood and its radius (Lemma 8), the probability of the neighborhood relates directly to its volume by the integration of the PDF (2b), if it exists.*
4) *The ball in Eq. (1) is defined as closed (not open), which implies a right-continuous NCDF. This is to follow the convention of many texts in defining univariate CDFs.*

**Definition 3** ($\widehat{NCDF}_{\|\cdot\|_p, X=x_i}$: sample NCDF). *Consider a dataset $\{x_i \mid x_i \sim i.i.d., i = 1, \ldots, n\}$ drawn from the r.v. $X$. A nonparametric estimator of $NCDF_{\|\cdot\|_p, X=x_i}$ is the sample NCDF defined as:*

$$\widehat{NCDF}_{\|\cdot\|_p, X=x_i}(v) = \frac{1}{n} \sum_j I_{\left(x_j \in \mathcal{N}_{\|\cdot\|_p}(x_i, V_{\|\cdot\|_p}^{-1}(v))\right)} \tag{3}$$

**Remark 3.**
1) *The sample NCDF converges pointwise to the population NCDF (Lemma 6).*
2) *The sample NCDF is a plot of the relative number of observations $\beta$, w.r.t. the total number of observations, in the neighborhood centered around the observation $x_i$ and has a volume $v$.*
3) *To draw $\widehat{NCDF}_{\|\cdot\|_p, X=x_i}(v)$, we need to compute $\|x_i - x_j\|_p$ $\forall j$, sort them, calculate the volume at each distinct distance, substitute in Eq. (3), and plot the points.*
4) *The resulting curve, therefore, will be a staircase with horizontal segments and vertical jumps. A horizontal segment $[v_1, v_2]$ represents the corresponding gap, in the feature space, between two empty neighborhoods, $\mathcal{N}_1$ and $\mathcal{N}_2$, of volumes $v_1$ and $v_2$, respectively. A vertical jump at $v_2$ has a value of $\Delta n/n$, where $\Delta n$ is the number of observations at the boundary of $\mathcal{N}_2$.*

**Definition 4** (Population and sample family curves). *Consider a r.v. $X$ and a drawn dataset $\{x_i \mid x_i \sim i.i.d., i = 1, \ldots, n\}$.*

*The population NCDF family of the r.v. is the set of curves* $NCDF_{\|\cdot\|_p, X} = \{NCDF_{\|\cdot\|_p, X=x} \mid x \in \text{support of } X\}$; *and the sample NCDF family of the dataset is the set of curves* $\overline{NCDF}_{\|\cdot\|_p, X} = \{\overline{NCDF}_{\|\cdot\|_p, X=x_i}, \ i = 1, \ldots, n\}$.

**Remark 4.**
1) *The complexity of generating* $\overline{NCDF}_{\|\cdot\|_p, X}$ *is obviously $n \times$ the complexity of generating* $\overline{NCDF}_{\|\cdot\|_p, X=x_i}$, *and the latter depends on the sorting algorithm (Section V-B).*
2) *The sample NCDF family should be interpreted as a set of characteristics of the dataset because each curve is a characteristic of its generating observation.*
3) *At particular proximity level $0 \leq \beta \leq 1$, a horizontal line (call it $\beta$-level), intersects with the $n$ curves at $n$ intercepts $v_i, \ i = 1, \ldots, n$; each value corresponds to an observation $x_i$. These $n$ intercepts are a sample of the population of volumes of $n$ neighborhoods of probability $\beta$. Therefore, if one of these intercepts has no close neighbors on this horizontal line, then this means that its corresponding observation is very different (anomalous) from the other observations. Said differently, other observations achieve the same $\beta$-level but at different volumes of neighborhoods.*
4) *An observation $x_i$ can look "anomalous" at some $\beta_1$-level but "normal" at some other $\beta_2$-level—if its NCDF curve has a point $(v_i, \beta_2)$ that has close neighbors $(v_j, \beta_2), \ \forall j \neq i$ on the horizontal line at $\beta_2$.*
5) *These characteristic NCDF curves are a function of the selected norm. Therefore, in principle, what does not seem anomalous under a particular norm may look anomalous under others.*

### B. Detection Algorithm and Visualization Approach

Algorithm 1 details UN-AVOIDS, our visualization and detection approach. The algorithm transforms the dataset from the feature space to the new NCDF space for both the visualization step, by rendering the curves to a plot, and the detection step, by assigning an anomalous score to each observation. The algorithm starts with normalizing each feature to $[0, 1]$; hence, the whole dataset will exist in the $d$-dimensional cube $[0, 1]^d$. The $n \times n$ matrix of distances between each pair of observations can be built under any selected $L^p$ norm; however, we chose an infinitesimal $p = 2^{-4}$, as explained in Remark 1 and Section V-C2. Because the volume of a particular neighborhood varies with the selected norm, normalization by the maximum volume under the selected norm is performed (Lemma 8). Finally, the sample NCDF family is generated and visualized (Section IV and its figures). Next, the analytical step starts by finding, for each NCDF, the $\beta$-level at which the NCDF has the largest horizontal gap that separates it from other NCDFs at the same level (Remark 4). Therefore, we cut the sample NCDF family at $l$ arbitrary levels, where each level cuts the family at $n$ intercepts, each intercept corresponds to a volume, and each volume encloses the same number of observations $\beta n$. At each $\beta$-level, out of the $l$ levels, the set of $n$ intercepts are compiled into the array `Inters`, and for a given NCDF, an anomaly score is assigned based on the distribution of these intercepts. The final outlier score for a given NCDF is then the

**Algorithm 1:** UN-AVOIDS

```
/*Initialize data and normalize features to [0,1]*/
Read X_{nxd}                        //the data matrix
for(j=1; j≤d; j++)    //normalizing each feature to [0,1]
    x_j = (x_j − x_{jmin})/(x_{jmax} − x_{jmin});

/*Construct NCDF space (a plot of n NCDF curves)*/
p = 2^{-4}                   //use a very small norm ||·||_p (Remark 1)
for(i=1; i<n; i++){       //build matrix V_{nxn} of ||·||_p distances
    for(j=i+1; j≤n, j++){
        V(i,j) = ||x_i−x_j||_p;
        V(j,i) = V(i,j);
    }
    Sort V(i,:);                 //ascendingly
    V(i,:) = V(i,:)/V(i,n);   //normalize by V_{max} (Lemma 8)
    PLT = Draw \overline{NCDF}_{||·||_p, X=x_i}//a plot of n NCDF curves
}

/*Visualization step: this is a human action (Section IV-C3
    provides an example). Analysts explore PLT and/or
    interact with it using a visualization aided anomaly
    detection (VAAD) software as in Discussion section.*/

/*Detection step: anomaly scores for the NCDF curves*/
Beta = [0.01,0.02,...,1];//dividing y-axis to l=100 β-levels
r = 0.01;                     //initialize r of ''fraction of gaps''
b = 0.05n;                    //initialize bins of ''histogram''
for(i=1; i ≤ n; i++){
    for(j=1; j≤l; i++){//assign anomaly score at each β-level
        Inters = getIntercepts(Beta[j], \overline{NCDF});//n×1 array
        switch (method){
            case "fraction of gaps":       //the fraction is r
                Gaps = |Inters − Inters[i]|;//n×1 array
                BetaScores[j] = getRthSmallest(r, Gaps);//1×1 array
            case "histogram":              //number of bins is b
                Counts = Hist(b, Inters);   //b×1 array (intercepts)
                for(int b', sum=0; b'=1; b'≤b)
                    //calculate relative probabilities in bins.
                    sum += (Counts[b']>Counts[i]) ? Counts[b'] : 0;
                BetaScores[j] = sum/n;
        }
    }
    Scores[i] = max(BetaScores); //n×1 array (anomaly scores)
}
```

maximum score across all the $l$ levels. It is quite important to emphasize that $l$ is not a complexity or tuning parameter of the algorithm. Rather, the higher the $l$, the more slicing of the NCDF space and the higher level of certainty that we will hit the $\beta$-level that achieves the highest separation. Because at two different $\beta$-levels that are very close to each other the intercepts will be almost identical, there is not much of a gain in precision, except for the consumed computational power, from setting a very large value of $l$.

In Algorithm 1, for the $i$-th NCDF, we introduced two methods for assigning an anomaly score at a particular $\beta$-level (`Beta[j]`). Each method assigns the score based on the relative location of the $i$-th intercept `Inters[i]` w.r.t. to the distribution of all intercepts `Inters`.

1) *"fraction of gaps"*: calculates the distances `Gaps` between `Inters[i]` and the other intercepts `Inters`, and then assigns the $\lceil nr \rceil$-th smallest distance, where $0 < r < 1$ is a relaxing parameter, as the anomaly score.

2) *"histogram"*: creates $b$ bins for the intercepts `Inters`, with the $i$-th intercept `Inters[i]` being centered in one of the bins. The count `Counts` of the intercepts in each bin is calculated, and then the anomaly score of the $i$-th NCDF is assigned as the count sum of all bins with a larger count than `Counts[i]`, which is the count of the bin of the $i^{\text{th}}$ intercept. It is very important to observe that the anomaly score by either method is normalized to $[0, 1]$, which is invariant with the data distribution. (For scrutiny, we use the "normalized score" to distinguish it from the "calibrated score" [46], where a score is both normalized and calibrated to a probability measure.)

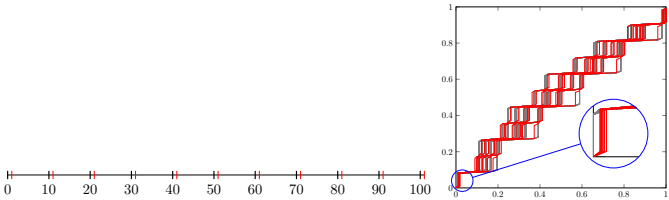

Figure 2: Left: Nine observations at $X = 0$ (black) and a local outlier (red); the pattern is repeated at 10 different locations $10, \cdots, 100$. Right: the corresponding sample NCDF, which is capable of distinguishing the outliers.

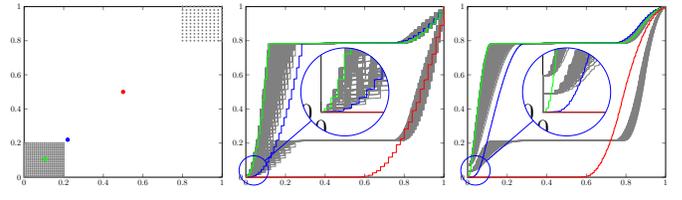

Figure 3: A simulated 2D dataset of two clusters and three possible outliers (left) and its sample NCDF with $p = \infty$ (middle) and $p = 2^{-4}$ (right). The NCDFs of the blue and red outliers are distinguishable; the NCDF of the green outlier is distinguishable only for $p = 2^{-4}$.

It is quite important to emphasize that the two methods provided above for assigning scores are just arbitrary and that there can be many other qualifying methods, not excluding the approaches from image processing (treating the NCDF space as an image), density estimation, among many others. We anticipate that many other methods can provide a good discriminating score, as well as the above two methods, because the NCDF space itself, the kernel of UN-AVOIDS, is what provided the separability among the anomalous and normal observations. Meanwhile, we shed more light on the two methods adopted above. We introduced the "fraction of gaps" methods because it is a very intuitive and natural candidate. The anomalous curve appears in the NCDF in isolation with respect to other curves, which makes the separating distance from the other curves a natural candidate for the anomalous score. However, to guarantee the scores always converge to a non-zero value, Lemma 9 shows that the distance should be measured between the curve of interest to the $K = nr$ nearest curve, where $0 < r < 1$. We introduced the "histogram" method, which is inspired by the concept of *probability of probability* [23], because it provides a different scoring mechanism than "fraction of gaps". It estimates the probability of each intercept within its locality; then it compares the probability of the intercept of interest to the other probabilities, assigning a score based on this relative probability comparison.

## IV. EXPERIMENTS

### A. Simulated Dataset in $\mathbf{R}$

We explain the capability of the NCDF to detect both local and global anomalies by giving a simple introductory example. Figure 2 illustrates a cluster of 10 observations belonging to $\mathbf{R}$: nine at $X = 0$ (black) and one at $X = 1$ (red), where it may be interpreted as an outlier. However, if this cluster is repeated along the real line at $X = 10i$, $i = 1, \ldots, 10$, the same red observation may not be considered anomalous anymore because it has a repeated pattern. Nevertheless, if we consider only its relative value to its local neighbor cluster, it still may be considered anomalous. Because this is an unsupervised problem, the decision indeed depends on the application. The visualization of the sample NCDF family (Figure 2) summarizes this configuration, because it illustrates two distinguishable clusters of curves, that correspond to the setup of the dataset.

### B. Simulated Datasets in $\mathbf{R}^d$

We constructed several simulated datasets in $d$-dimensions, $d = 2^{d'}, d' = 1, \ldots, 9$; each is normalized to fit the $d$-dimensional unit cube $[0, 1]^d$ and denoted as $\mathcal{D}_d$. The dataset $\mathcal{D}_2$ is shown in Figure 3, where it comprises two clusters with two different densities, in addition to three observations that are nominated to be outliers. Cluster $\mathcal{C}_1 = \{(0.01i_1, 0.01i_2) \mid i_1, i_2 = 0, \ldots, 20\}$ has $21^2$ observations, cluster $\mathcal{C}_2 = \{(0.8 + 0.02i_1, 0.8 + 0.02i_2) \mid i_1, i_2 = 0, \ldots, 10\}$ has $11^2$ observations, and the three outliers are $(0.105, 0.105)$ in green (observe that it does not belong to the set $\mathcal{C}_1$), $(0.22, 0.22)$ in blue, and $(0.5, 0.5)$ in red. It is clear from the figure that the three outliers have distinguishable curves in the NCDF space. The green outlier is the most challenging one to observe. The reason for this is that although it is not exactly following the same geometric lattice pattern of the cluster, its distancing pattern is very similar. However, it is remarkable that it is distinguishable in the NCDF space under the norm $\|\cdot\|_{2^{-4}}$.

The setup of the datasets in higher dimensions is similar, except that we keep the total number of observations fixed at $11^2 + 21^2$ (otherwise, it would be $11^d + 21^d$, which would be massive for higher $d$), in addition to the three outliers. For $\mathcal{C}_1$, we sample $11^2$ observations uniformly from the grid $\{(0.01i_1, \ldots, 0.01i_d) \mid i_{d'} = 0, \ldots, 20\}$. The same setup is followed for $\mathcal{C}_2$. The three outliers follow the same pattern of those in the 2D dataset and are injected at $(a, \ldots, a)$, $a = 0.105, 0.22, 0.5$. Figure 4 illustrates the NCDF space $\forall d = 2^{d'}, d' = 1, \ldots, 9$ and $\forall p = 2^{p'}, p' = -4, \cdots, 2, \infty$. It is obvious that the three outliers are distinguishable in the NCDF space for almost all values of $d$ and $p$. They are more distinguishable for a higher $d$ and smaller $p$.

### C. Real Datasets

#### 1) Setup: We assessed UN-AVOIDS using two recently published real datasets that contain a wide variety of malicious

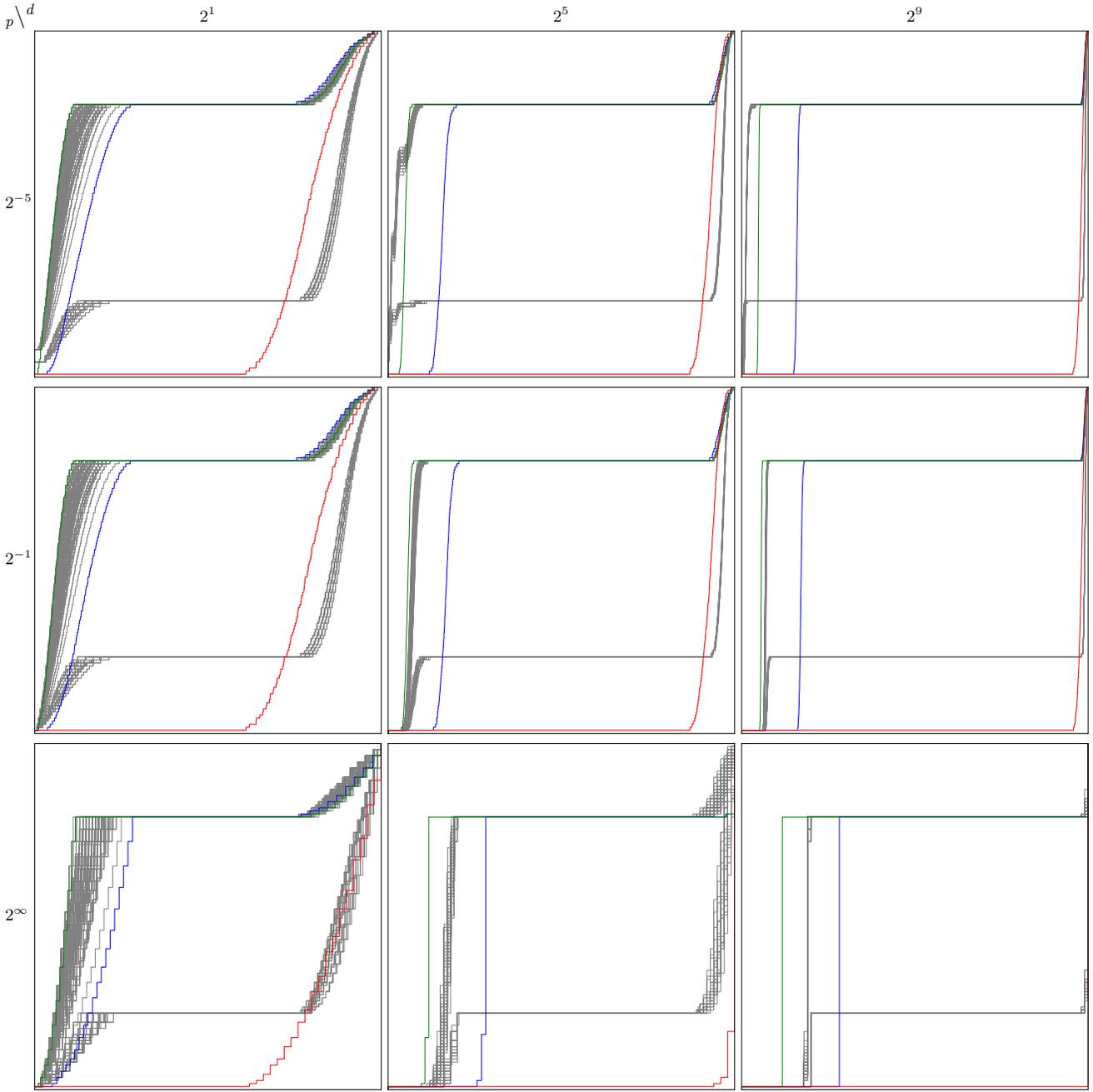

Figure 4: The $\widehat{NCDF}_{\|\cdot\|_p, X}$ for $p = 2^m, m \in \{-5, -1, \infty\}$ (rows), and $d = 2^m, \; m \in \{1, 5, 9\}$ (columns) for a toy dataset (the first column is an expanded version of Figure 3, for several values of $p$).

attacks. `CICIDS2017` was published in [47], and follows the 11-criterion framework proposed in [48]: complete network configuration, complete traffic, labeled, complete interaction, complete capture, available protocols, attack diversity, anonymity, heterogeneity, feature set, and metadata. None of the datasets that were publicly available prior to `CICIDS2017` featured all 11 criteria [47, 48]. After removing observations with invalid values, `CICIDS2017` comprises 2,827,595 observations, 556,541 of which are malicious. Each observation has 84 features, including a timestamp. We used only the

76 features that were extracted using the CICFlowMeter of the Canadian Institute for Cybersecurity[1] (CIC). The dataset contains a variety of attacks, such as Brute Force (FTP-Patator and SSH-Patator), DoS (Slowloris, Slowhttptest, Hulk, and GoldenEye), DDoS LOIT, Heartbleed, Web-based (Brute Force, XSS, and SQL Injection), Infiltration (Dropbox Download and Cool disk – MAC), Meta exploit Win Vista, Port Scan (with and without Nmap), and Botnet ARES.

---

`CIRA-CIC-DoHBrw-2020` was outlined in [49] and also was provided by the CIC[2]. The dataset contains DNS over HTTPS (DoH) traffic, which is a newer protocol designed to secure DNS traffic to combat DNS manipulation during the carrying out of man-in-the-middle attacks. We used only the malicious DoH traffic and benign non-DoH traffic, and abandoned the benign DoH traffic because it only contributes to 1.7% of the dataset size and to have consistency with the other dataset. After removing observations with invalid values, `CIRA-CIC-DoHBrw-2020` comprises 1,139,362 observations, 249,836 of which are malicious. After removing IP/Port and timestamp info, only 29 features remained. The dataset is more homogeneous, with a single type of attack (man-in-the-middle), where the malicious traffic is created with tunneling tools. DNSCat2 and dns2tcp were used to create the false DNS query results, which, in practice, would misdirect the web client, resulting in sending data to malicious servers, and hence leading to it being stolen.

We used these two real datasets under a wide range of configuration parameters to compare the performance of five algorithms: UN-AVOIDS (with the two versions of the scoring methods of Algorithm 1) and the three algorithms LOF, IF, and FABOD (Section II-A). We chose these three algorithms for comparison against UN-AVOIDS for several reasons. (1) They belong to three different families of outlier detection algorithms: density based, angled based, and classification, respectively. (2) These methods showed excellent performance in [50], where they were compared with nine other outlier detection algorithms across five datasets. The authors reported that, in terms of the area under the ROC curve (AUC), the best two algorithms were ABOD and its faster implementation, FABOD, respectively. IF performed favorably and achieved the highest F1 score, whereas LOF achieved a very high precision. (3) These three algorithms are popular and widely available in important libraries, such as `scikit` [31] and PyOD [51].

Each dataset was randomly shuffled; and to experiment with different values of prevalence, we sampled the malicious observations at a prevalence rate, $\pi = 2^m/1000, m = 0, \cdots, 5$, and then discarded the rest. Next, the whole dataset of size $n$ was partitioned into $n/ws$ windows, where the window size $ws = 2^m \times 100, m = 2, \cdots, 5$. Because the malicious observations were not uniformly distributed across the dataset, their number per window is a r.v. For each value of $ws$ and $\pi$, we aggregated the anomaly scores assigned by a particular algorithm to all observations in a given window, then calculated the AUC using the Mann-Whitney statistic before averaging the AUCs across all windows. This AUC is the outcome of a single experiment out of a total of 1008 experiments, which are the cross-product of the above parameter configurations. Because AUC is prevalence independent, we always advocate for using it to assess classification rules rather than other prevalence- and threshold-dependent measures, such as accuracy, $F_\beta$ measure, and so forth (for a more detailed account of the AUC, the reader is referred to [52]). For UN-AVOIDS, we experimented with $p = 2^m, m = -5, \cdots, 2, \infty$, and set $l$, $r$, and

$b$ (Algorithm 1) to 100, 0.01, and $0.05n$, respectively. Although, as explained in Remark 1 and Section V-C2, we should use an infinitesimal value of $p$, we experimented with these several values just for emphasizing this fact and for demonstration. For the other three algorithms, we used their implementation in [31, 51] with the following parameter settings: for LOF, we used the technique suggested by the authors in [30], where we take the maximum outlier score across a set of `MinPts`; for IF, we used the default value of 100 for the number of trees; and for FABOD, we set the number of neighbors to 160.

*2) Detection:* The results of the above configurations are illustrated in Figure 5, which comprises a total of 72 subfigures: 2 (datasets) × 9 (values of $p$) ×4 (values of $ws$). Each subfigure is a plot of the AUC of the five algorithms vs. the prevalence index $m$, at particular $p$ and $ws$. Since LOF, IF, and FABOD are not a function of $p$, their curves are identical for all subfigures of a particular dataset and $ws$ (column); however, they are included in these subfigures for the sake of comparison with UN-AVOIDS. From the figure, it is clear that UN-AVOIDS is almost an overall winner at $p = 2^{-4} (<< 1)$ for both datasets and all values of $\pi$ and $ws$, with probably very few exceptions. Depending on $ws$, $\pi$, and the dataset, the AUC gain varied between -0.028 (= 0.7167 − 0.7451, the difference between UN-AVOIDS with "histogram" at $ws = 400, m = 2$, and FABOD, respectively, on `CIRA-CIC-DoHBrw-2020`), and 0.125 (= 0.8030 − 0.6781, the difference between UN-AVOIDS with "histogram" at $ws = 3200, m = 6$, and FABOD, respectively, on `CIRA-CIC-DoHBrw-2020`). The behavior with $p$ complies with both the results of the simulated dataset (Figure 4), and the intuition behind the very small $p$ neighborhood and the capability of its hunting spikes to grow without consuming the Euclidean space (Remark 1 and Section V-C2). Decreasing $p$ more greedily deteriorates the performance. The interpretation can be anticipated from the mathematical fact that $\lim_{p \to 0} \mathcal{N}_{\|\cdot\|_p, A}(x_0, \epsilon)$, $\forall \epsilon$ shrinks significantly, and gets squashed to the axes. The shrunk neighborhood is a set of zero probability measure for continuous features. Only some of the discrete features may still live in such a degenerate space; however, they may not contain enough information on the anomalousness because they are a subset of the whole feature set. The effect of increasing $p$ is almost dataset dependent, with a little decrease with higher $p$, and significant decrease at $p = \infty$. The effect of $ws$ is in favor of UN-AVOIDS more than other algorithms. In addition, the performance of UN-AVOIDS is almost stable with $\pi$, which is in contrast to the other three algorithms that deteriorate with higher $\pi$. Finally, although UN-AVOIDS is an overall winner, its winning scoring method is different for each dataset: for `CICIDS-2017` and `CIRA-CIC-DoHBrw-2020` it is "fraction of gaps" and "histogram", respectively. This motivates more attention to designing a more consistent scoring method, as discussed in Sections III-B and V-C3

Next, we compare UN-AVOIDS with the basic baseline methods, namely $K$-NN and $K$-means, with and without principal component analysis (PCA), on the same two real datasets. Figure 6 illustrates that at all examined values of $K$ (= 1,3,5,7,9,19,29), UN-AVOIDS outperforms both methods on `CIRA-CIC-DoHBrw-2020`. On `CICIDS-2017`, $K$-NN slightly outperforms UN-AVOIDS only at $K = 1, 3$; whereas $K$-means outperforms it at $K = 3, 5, 7, 9, 19$. Two important comments are in order. First, even though this is a case of failure of UN-AVOIDS, there is no way to anticipate the winning value of $K$ for each dataset. Second, outperforming at some values of $K$ on this dataset means that both baseline methods outperformed the other methods (IF, LOF, and FABOD) as well. This is very unlikely per the comparative study [50], which indicated that both methods achieved very low AUC values, and ranked the fifth and above among other methods. Nevertheless, this supports the "no-over-all-winner" principle, where "any discrimination rule can have an arbitrarily bad probability of error for finite sample size" [53].

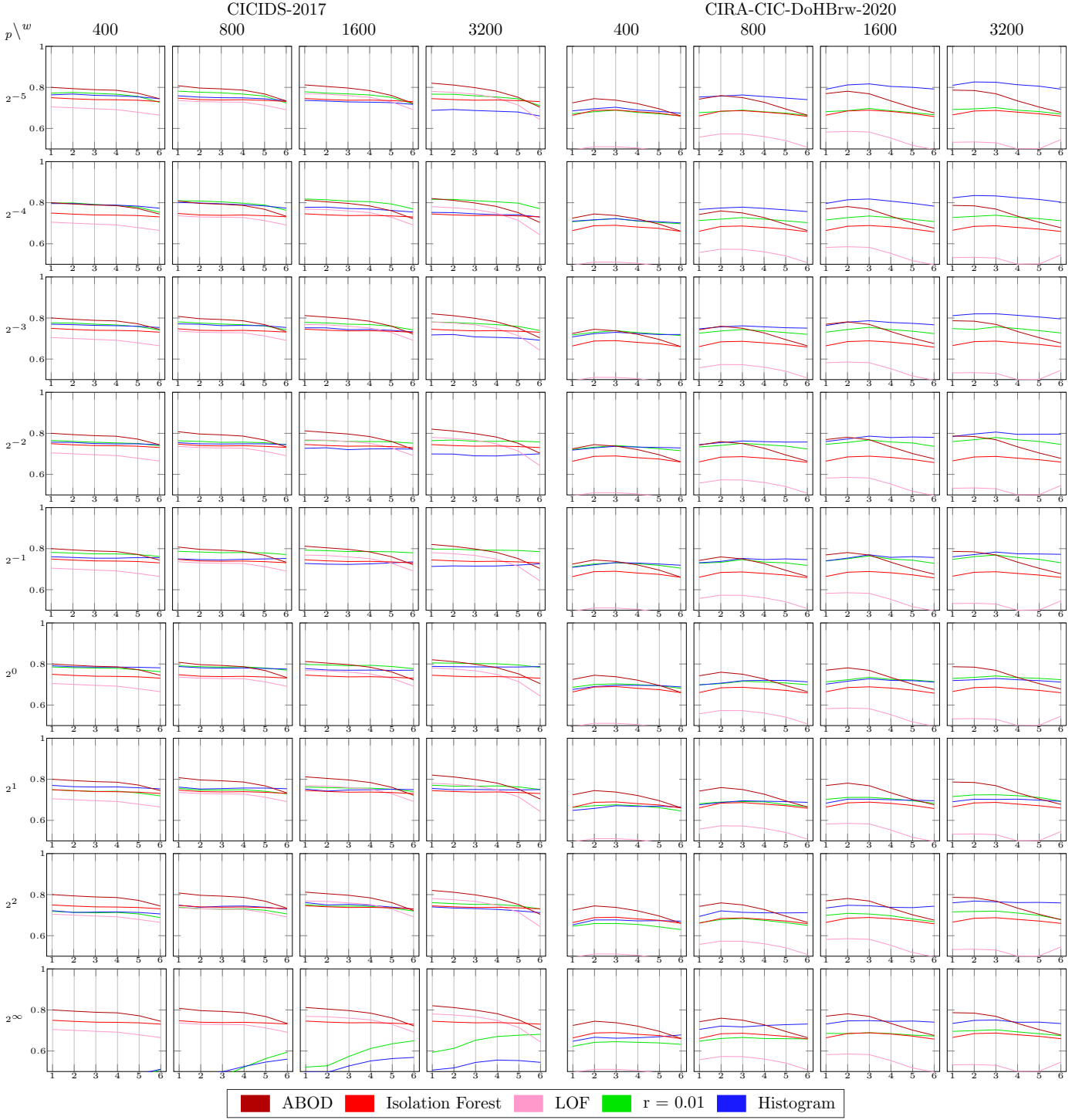

Figure 5: AUC of 1,004 experiments (the cross-product of): 2 datasets (`CICIDS2017` and `CIRA-CIC-DoHBrw-2020`), 5 algorithms (LOF, IF, FABOD, and 2 scoring methods of UN-AVOIDS × 9 values of $p$ (rows)), 4 values of $ws$ (columns), and 6 values of $\pi$ ($2^m/1000, m = 0, \cdots, 5$, on $x$-axes). At $p = 2^{-4}$, UN-AVOIDS is almost the overall winner: "fraction of gaps" and "histogram" wins for the two datasets, respectively.

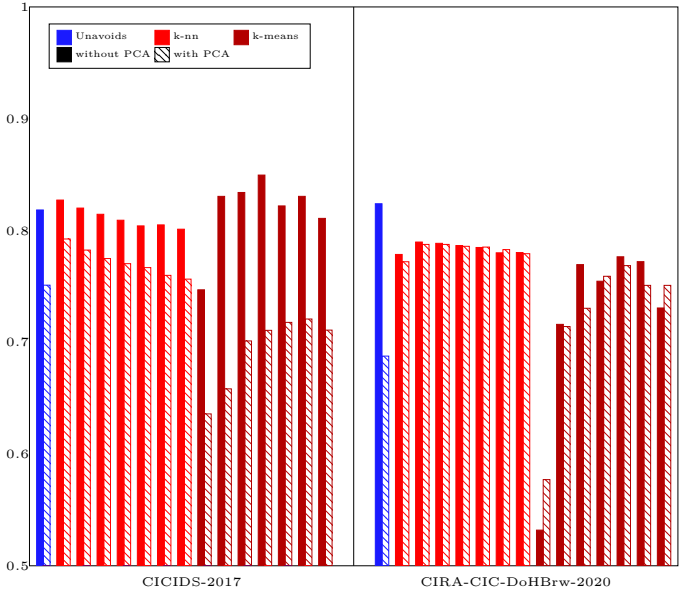

Figure 6: AUC comparison between UN-AVOIDS, KNN and Kmeans, with and without PCA, on the two real datasets. The PCA is not helpful. Only on CICIDS2017, and for some values of $K$, the $K$-NN and $K$-means slightly outperform UN-AVOIDS. The best value of $K$ is not consistent and requires supervised learning.

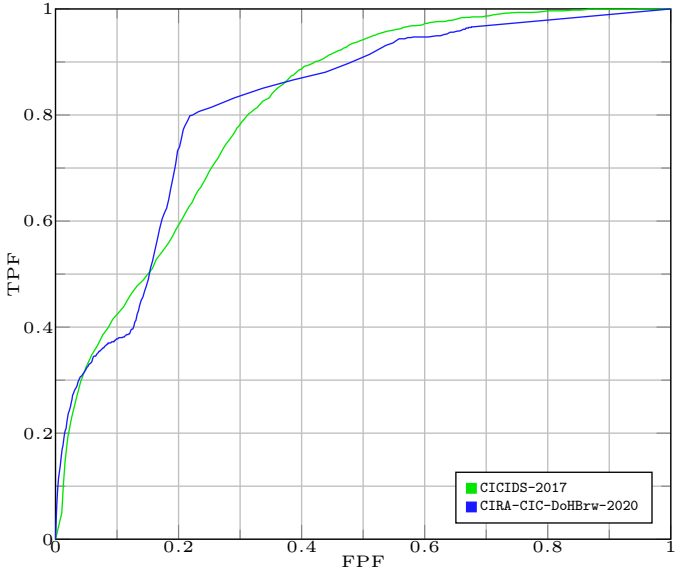

Figure 7: The ROC of UN-AVOIDS with $p = -4$ and with the highest scoring window size on both real datasets.

Next, we present the ROC curve itself, rather than its summary of measure AUC, of UN-AVOIDS on the two real datasets for $p = -4$, $ws = 400$ (Figure 7). The ROC curve is obtained by aggregating the scores of each window, and compiling the scores of the whole dataset. Instead of calculating the AUC using the Mann-Whitney statistic, a decision threshold $\theta$ is chosen, and then the TP fraction (TPF = fraction of anomalous observations with scores $> \theta$) and FP fraction (FPF = fraction of normal observations with scores $> \theta$) are calculated. This constitutes a single operating point $(FPF, TPF)$ on the ROC curve. The threshold $\theta$ is varied (from the minimum to maximum possible score), generating several operating points, and the ROC curve is generated. The area under the generated ROC curve (AUC) is identical to the AUC calculated from the Mann-Whitney statistic of the compiled scores [54]. We emphasize that the threshold is not a tuning parameter; rather, this is the case for any classifier providing a decision score. Then, applying a decision threshold results in a binary decision rule with an FPF and TPF of a particular operating point $(FPF, TPF)$ on the classifier's ROC curve. It is obvious that designing a decision rule that provides a score, like UN-AVOIDS, can provide a continuum of operating points on the ROC curve by playing with the threshold $\theta$ and is more general and flexible for applications than the special case of providing a binary decision that has a single operating point on the ROC curve. Later, a practitioner can choose the threshold that achieves a particular operating point based on the type of problem or application.

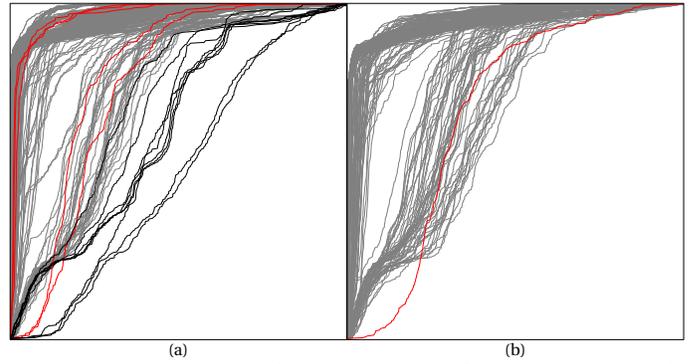

Figure 8: The NCDFs of two windows of CICIDS2017, with $ws = 400$, $p = 2^{-4}$, and $\pi = 0.004$. By thresholding the scores, whose AUC = 0.86 and 1, at 0.0564 and 0.0925, the TN (gray), FN (red), FP (bold gray), and TP (bold red) are 386, 3, 9, 2 (i.e., 97.7%, 60%, 2.3%, 40%), and 399, 0, 0, 1 (i.e., 100%, 0%, 0%, 100%), respectively.

*3) Visualization:* We illustrate the visualization capacity of UN-AVOIDS at $p = 2^{-4}$ with two example windows from CICIDS2017, with $ws = 400$. The scores provided by UN-AVOIDS for each observation were obtained via the "fraction of gaps" method. The two windows comprised five and one malicious observations, exhibited a moderate and perfect AUC = 0.86 and 1, and produced the sample NCDF families in Figures 8a, and 8b, respectively. After using a decision threshold of 0.0564 and 0.0925, the TN (gray), FN (red), FP (bold gray), TP (bold red) can be easily counted from the figure as 386, 3, 9, 2 (i.e., 97.7%, 60%, 2.3%, 40%), and 399, 0, 0, 1 (i.e., 100%, 0%, 0%, 100%), respectively. The visual performance of the family in Figure 8b is impressive. It is interesting to note that the only TP has a *DoS Hulk* attack type. For the family in Figure 8a, the TP is obvious, and the FP is very small and indeed deserves investigation from the administrator, and therefore we cannot consider it FP yet. The FN curves are not distinguishable and are hiding within the TN curves; however, we have to recall that this is a synthesized dataset, and what was labeled as malicious needs not be a real anomaly in a practical setup.

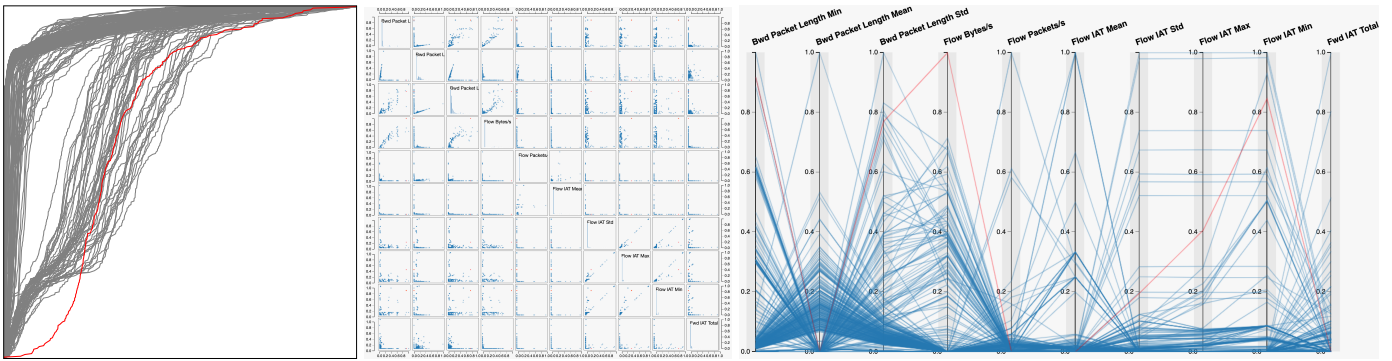

Figure 9: Brushing and indirectly linking the NCDF of Figure 8b and both the ⊞-mat and ‖-cords of the 10 features of the same windows' observations. The 2D NCDF space is remarkably more capable of distinguishing the outlier.

For more emphasis on the visualization power of the NCDF space, Figure 9 illustrates the sample NCDF family of Figure 8b and both the scatter plot matrix (⊞-mat) and the ‖-cords of the first 10 features of the 400 observations of the same window; both figures were produced and exported by the Data Visualization Platform[3] (DVP) [55]. Several important remarks are in order: (1) The space limitation makes it very difficult to use either the ⊞-mat or ‖-cords for exploring the whole set of 76 features of the data. Even when rendering on a huge screen, it is extremely difficult to mutually explore 2850 (=$^{76}C_2$) plots in the ⊞-mat, and 76 axes in the ‖-cords. (2) It is quite impressive that the 2D NCDF space—though not a lower subspace for data representation (Section III)—clearly distinguishes the outlier. (3) After selecting and brushing the outlier that is obvious in the NCDF space, we used its observation ID to brush the other two plots on DVP before exporting them. Putting the three plots side by side provides more intuition to the system administrator, in particular, or the data analyst, in general. The NCDF shows "who" is anomalous, while the other linked and brushed plots show "why". This is because the information they convey comes directly from the original feature space, which resembles the physics of the problem, and not from a transformed space; hence, this information can be parsed by the analyst. In Section V-D, we discuss the need for new visualization software to harvest the full utilization of both the detection and visualization capabilities of UN-AVOIDS.

## V. Discussion

### A. Computational Power

All experiments on the real datasets were carried out on the Compute Canada Cedar Cluster[4] (CCCC), which has a massive computational power. However, conducting the experiments for the entire project was very computationally expensive and consumed several months of continuous computations. We demonstrate this by explaining only the production of Figure 5, aside from other trial-and-error or experimental configurations. Because the experiments were conducted for both datasets by finding the NCDFs for all observations, determining their outlier scores, under all parameters, for a given window, and then repeating over windows, we used job arrays. This means that we scheduled multiple independent jobs (scripts), with all jobs being run in parallel, each with 4 GBs of memory and 4 CPUs. Jobs were executed at different speeds because the windows had different sizes. Respectively, `CICIDS2017` and `CIRA-CIC-DoHBrw-2020` comprised 7,068 and 2,848 jobs that took approximately 114,379 h and 46,102 h, which sum up to 160,481 h (=18 years) of serial computing time. Thanks to CCCC, where we were able to run, on average, 500 jobs in parallel, this reduced the computational time to 13 days.

### B. Complexity and Computational Time

UN-AVOIDS' computational complexity can be decomposed into two components: generating the NCDF and calculating the anomaly scores. Generating the NCDF must be carried out, first, for the whole dataset, which is $O(dn^2)$, because it requires calculating $\|\cdot\|_p$ between each pair of its $n$ observations. This implies that UN-AVOIDS prefers a small window size in real-time applications. Next, testing a single future observation is clearly $O(dn)$. We succeeded in speeding up the computational time by taking advantage of parallelization and the use of optimized libraries for matrix multiplication, such as `numpy`.

The complexity of calculating the scores depends on the scoring method (Algorithm 1). For the "fraction of gaps" method, it is necessary to sort the $n$ intercepts at each $\beta$-level to find the nearest subset of them with size $rn$. This step was done for all levels a priori, which is $O(ln \log(n))$. Afterwards, then at a particular $\beta$-level, for the $i^{th}$ intercept, we know that the nearest $rn$ intercepts are laying within the $2rn$ nearest indices in the list of sorted intercepts. We take the differences between the $i^{th}$ intercept and these $2rn$ nearest indices, sort them, and then select the $rn^{th}$ smallest difference, which is $O(n \log(n))$; and because it is pursued for all observations at all levels, it is as demanding as $O(ln^2 \log(n))$, which dominates the initial sorting step. For the "histogram" method, it is not necessary to sort the intercepts because binning each intercept can be done in $O(1)$ using the bin boundaries. However, $n$ sets of bin boundaries are computed for each $\beta$-level in order to centre each intercept in its bin when computing its outlier score, so for each $\beta$-level, we must bin the $n$ intercepts $n$ times. Therefore, the overall complexity is $O(ln^2)$. Although the quadratic complexity of UN-AVOIDS can be regarded as a limitation, a comparison with other algorithms, which we do next, reveals that some of them have the same complexity.

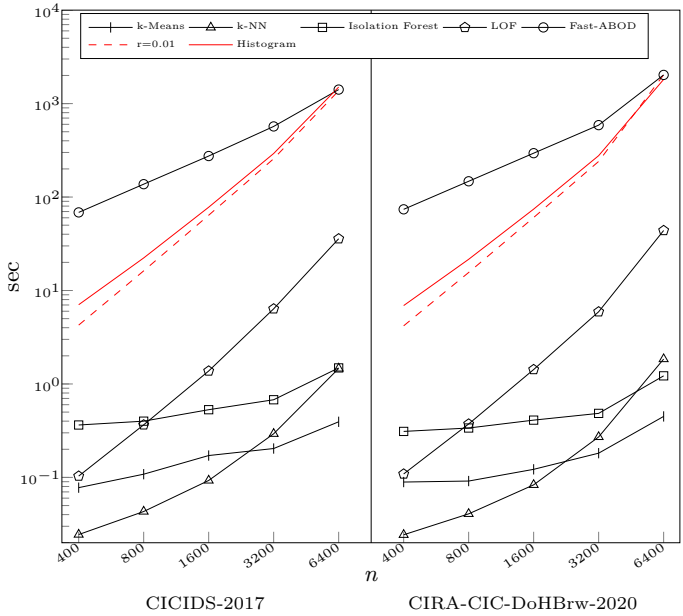

Figure 10: Comparative execution time (on a $\log-\log$ scale) for the different anomaly detection algorithms that processed the two real datasets, at different window sizes.

*1) LOF:* [30] reported $O(n^2)$ and $O(n)$ time complexities for the first and second step of the algorithm, respectively. The first step requires finding the $K$ nearest neighbors of each observation. Because the first step dominates, and requires computing the distances of $d$-dimensional vectors, the final complexity should be $O(dn^2)$.

*2) FABOD:* [33] reported a time complexity of $O(n^2 + nK^2)$. The $n^2$ term comes from computing the $K$ nearest neighbors of each observation. The $nK^2$ rises because, for each observation, the angle between every pair of its $K$ neighbors is computed. Because both steps require taking the dot product, the complexity becomes $O(dn^2 + dnK^2)$.

*3) IF:* [56] reported a complexity of $O(t\psi^2)$, where $t$ is the number of trees and $\psi$ is the "sub-sampling size", which is 256 in our case, the default value for the chosen implementation in PyOD. The default of this implementation samples $d$ split values per node. There are $\psi$ nodes per tree; and during the creation of each node, all $d$ feature values are involved in a comparison for each of the $\psi$ samples. Therefore, the resultant complexity is $O(dt\psi^2)$.

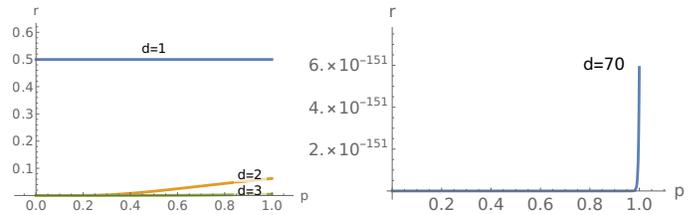

Figure 11: Vanishing neighborhood in high dimensions under $p < 1$ and hitting machine precision. The ratio $V_{\|\cdot\|_p}(\epsilon)/V_{max}$ (Lemma 8), with $\epsilon = 0.5$ at $d = 1, 2, 3$ (left) and $d = 70$ (right).

*4) K-means:* scikit's implementation uses Elkan's algorithm [57], whose authors reported a complexity of $O(nKi)$, where $K$ is the number of centroids and $i$ is the number of iterations. For each iteration, $O(nK)$ comes from finding the lower bound $l(x, c)$ for each observation $x$ and mean $c$, which requires finding the distance from each observation to each mean. Therefore, the final complexity is $O(dnki)$.

*5) K-nn:* The distance between each pair of observations must be computed, which gives a complexity of $O(dn^2)$.

It is informative to investigate the actual execution time of each of these algorithms on the two real datasets and at different window sizes $n$ (Figure 10). This is the total execution time, which comprises: the construction time, e.g., building the NCDF space in the case of UN-AVOIDS, and the testing time required to assign the final score to a single testing observation. The figure uses a $\log-\log$ scale for a clearer comparison among the different powers and scales of $n$, so that $t = cn^p$ becomes $\log t = \log c + p \log n$, which is linear. All methods exhibit almost linear behavior with different slopes ($p$) and different intercepts ($\log c$). Therefore, all quadratic algorithms have almost a double slope than the linear ones. It is clear that most methods, except FABOD, win over UN-AVOIDS in the computational aspect. This strongly motivates the implementation of a faster version of NCDF construction, without sacrificing much accuracy (Section V-C), which decreases the slope of its line in the figure. Another remedy could be designing a new scoring method (Sections III-B and V-C3) that does not scan the $l$ different values of $\beta$-levels, which, despite of keeping the complexity unchanged, can bring UN-AVOIDS' line in Figure 10 below $K$-NN and LOF by eliminating the $\log l$ intercept.

### C. Future Work: Analytical and Algorithmic

The first avenue that we foresee opening up for UN-AVOIDS' development concerns its analytical and algorithmic approach, where the contributions can either provide new features or overcome current limitations, as explained below.

*1) Space Equivalence:* An extension to Lemma 7 is the following conjecture: "Both $NCDF_{\|\cdot\|_p, X=x}$ and $F_X$ are equivalent $\forall p, d$". It is of both theoretical interest, and probable practical implications to future applications, to prove or disprove this conjecture. However, we think that it is a very challenging mathematical problem that requires a rigorous topology, which is out of the scope of the present article.

*2) Hunting Spikes ($p << 1$) and Machine Precision:* The argument of hunting spikes when $p << 1$, introduced in Remark 1, concerns the curse of dimensionality and the topology of the space. We do not consider $p$ to be a tuning parameter because we only consider infinitesimal $p << 1$ per the argument. However, decreasing $p$ more greedily below the value $2^{-4}$ deteriorated the performance for both datasets ($d = 74, 29$) because it shrank the neighborhood $\mathcal{N}_{\|\cdot\|_p, A}(x_0, \epsilon)$, $\forall \epsilon$ to a region of almost zero probability measure, except for discrete variables, which may carry no information on the anomalousness (Section IV-C2). Here, we shed more light on this phenomenon and relate it to the machine precision, as opposed to algorithm complexity and tuning. Figure 11 illustrates the ratio $r = V_{\|\cdot\|_p}(\epsilon)/V_{max}$, the volume of the neighborhood $\mathcal{N}_{\|\cdot\|_p}(x_i, \epsilon)$ relative to the maximum volume of the dataset hypercube, versus $p$ at different values of dimensionality $d$. The neighborhood vanishes rapidly with increasing $d$ and/or decreasing $p$. For example, at $d = 70$, it starts reaching the order of magnitude of $10^{-150}$ only at $p = 1$. This is very challenging for machine precision and for the probability of finding any data in this squashed space.

*3) Scoring Method:* Equally important, a more elaborate scoring function, than "fraction of gaps" and "histogram" (Algorithm 1), is needed to be a consistent winner over different datasets and to assign lower score values to normal observations for increasing the AUC and decreasing the FBF at arbitrary thresholds. As mentioned in Section III-B, these methods can stem from several approaches, e.g., image processing, density estimation, among many others. We have already designed several other scoring methods, such as averaging across multiple $r$, and averaging all the distances from a given intercept to all other intercepts on a $\beta$-level. Some of these methods won over the two adopted methods but for only a few configurations and with no consistency.

*4) Data Rotation:* From Lemma 5, some important properties are preserved for the neighborhood of Eq. (1) under any transformation $A$. One of the candidate transformations is rotation, which preserves the relative geometry of the space. A rigorous mathematical analysis and a comprehensive empirical study are needed to study the effect of dataset rotation on the performance of UN-AVOIDS. It can be intuitively anticipated that under appropriate rotations, some outliers will be brought to the scope of the hunting spikes of the $L^p$ norm, $p << 1$. Although we tried a few experiments on rotation using random directions in $\mathbf{R}^d$ with no success, we did not have the luxury of designing a comprehensive set of experiments, because the computational power of CCCC's resources allotted to us was fully consumed for several continuous months as explained in Section V-A.

*5) Time Complexity:* Because building the NCDF space is $\mathcal{O}(dn^2)$, it will not be efficient for detecting anomalies in the whole dataset, where window slicing is not allowed (as in our experiments for network data). One possible remedy is via parallelization, as was pursued in our experiments. A second alternative is investigating the possibility of designing a fast version of the algorithm without sacrificing much accuracy, as explained in Section V-B.

*6) Handling Categorical Variables:* UN-AVOIDS is designed to work in a metric space, in particular $\mathbf{R}^d$; therefore, it cannot handle categorical variables directly in their raw format. However, one direct remedy is available via feature encoding (a wide range of literature is reviewed in Section II-A.)

*7) Handling Unstructured Data, e.g., Images:* An image of $m \times n$ pixels is a vector in $\mathbf{R}^{m \times n}$, which has a huge dimension. Moreover, the anomalousness may be resembled at deeper composed levels of data representation than the plain pixels, e.g., semantics of image components and objects. Therefore, using the plain pixel values for anomaly detection may not be prudent. A straightforward remedy is using deep autoencoders, a well-known variant of deep neural networks (DNN), that finds a reduced latent sub-space $\mathbf{R}^{d' << m \times n}$, in which each image is represented under some loss function minimization. Consequently, UN-AVOIDS can be applied to the new representation in $\mathbf{R}^{d'}$.

### D. Future Work: Visualization Software

The second avenue of UN-AVOIDS' concerns utilizing its visualization capabilities. It is of great importance to design data visualization software that aids the analyst in detecting outliers through exploration, as opposed to explanation (Section I). Hence, we coin the expression *visualization aided anomaly detection* (VAAD), which should provide two major sets of functionality. (1) VAAD should provide the visualization component of UN-AVOIDS, i.e., the NCDF space, along with several other plots, such as ⊞-mat, ∥-cords, multi-dimensional scaling (MDS), grand tour in high dimensions, and many others, in a single interactive visualization environment, where linking and brushing are provided. In such an environment, creating Figure 9 will require a few mouse-clicks. In addition, the user will be able to pursue data exploration to understand "why" anomalies are so. (2) VAAD should provide the analytical component of UN-AVOIDS, i.e., the scoring methods ("fraction of gaps", "histogram", among possible others), working in the back engine to provide anomaly scores of the visualized observations, along with an aggressiveness level—a threshold that controls the sensitivity and specificity (recall Figure 8). In this manner, selecting a particular score range, or setting a hard-decision threshold value, will brush the corresponding NCDF curves and the corresponding observations in the other plots, and vise versa. However, to objectively quantify its visualization utility for analysts—as we objectively quantified the power of its detection algorithm—UN-AVOIDS should undergo an assessment akin to the assessment of computer-aided detection (CAD) software. In this assessment, radiologists read mammograms without, then, with looking at the CAD's output, providing a decision after each reading. The hypothesis that the software improved the detection is tested at the required level of significance.

From the software engineering perspective, designing a VAAD can start from scratch or by building on other existing generic and flexible platforms, such as DVP [55]. It is quite important to emphasize that the utility of this VAAD is not exclusive to cybersecurity. It is applicable to any domain that generates data, such as rare disease detection and fraud detection of financial transactions, where the positive class is always scarce and the negative class is abundant, a setup where supervised learning is a challenge.

## VI. Conclusion

The present article considered the problem of visualization and detection of anomalies (outliers). From the literature review, we concluded that all approaches, in many literatures in general and in cybersecurity in particular, treated the two problems separately and provided a solution for only one problem at a time. In addition, the visualization approaches were introduced for explaining the output of a detection algorithm, not for providing a standalone visual exploration method. We introduced UN-AVOIDS to fill this gap in the literature and to provide both approaches in one coherent framework, by converting data to the NCDF space, in which anomalies have distinguishable curves. In this space, both the detection algorithm and the analyst see the same information, and hence, both the visualization power of the brain (human process) and the auto-detection (algorithmic process) can work simultaneously and cooperatively.

To assess UN-AVOIDS, we conducted several experiments on both a simulated and two real datasets from cybersecurity that were recently published and comprised several types of attacks. The assessment on the real datasets was compared with three of the most powerful anomaly detection methods: LOF, IF, and FABOD. UN-AVOIDS was almost an overall winner at $\|\cdot\|_{2^{-4}}$. We illustrated the visual exploration power of UN-AVOIDS by comparing the NCDF space to ⊞-mat and ∥-cords plots. The power of the new NCDF space stems from two facts: first, the anomalies become very distinguishable, and second, because it is a 2D space, the information is easily parsed when compared with the $\binom{d}{2}$ subfigures in ⊞-mat, or the $d$ parallel axes in ∥-cords.

We concluded the article with a discussion on our vision of possible avenues, that are open for elaboration and enhancements, in both theoretical and practical aspects. One important possibility to aid analysts is to develop VAAD software that brings UN-AVOIDS' visualization and detection, along with other conventional visualization methods, into one interactive environment.

## VII. Appendix: On the Mathematical Properties of NCDF

The full theory of the new NCDF space, on which UN-AVOIDS is founded, still needs more development. In this section, we provide some of its mathematical properties.

**Lemma 5.** *The neighborhood* (1) *is convex for every $d$, square matrix $A$ (even if neither symmetric nor positive definite), and any norm $\|\cdot\|$; and for $\|\cdot\|_p$, it is convex only for $1 \le p$.*

**Proof.** First, $\|A^{-1}(x - x_c)\| = (x - x_c)'A^{-1'}A^{-1}(x - x_c)$; and because $A^{-1'}A^{-1}$ is always symmetric and p.d., this drops any constraint on $A$. Next, $\mathcal{N} = \{x \mid \|A^{-1}(x-x_c)\| \le \epsilon\} \equiv \{x = x_c + Au \mid \|u\| \le 1\}$. Then, $\forall x_1, x_2 \in \mathcal{N}$, $x_1 = x_c + Au_1, \|u_1\| \le 1$, $x_2 = x_c + Au_2, \|u_2\| \le 1$. Then, for $0 \le \theta \le 1$, a convex combination $x = \theta x_1 + (1-\theta) x_2 = x_c + Au$, where $\|u\| = \|(\theta u_1 + (1-\theta) u_2)\| \le \theta \|u_1\| + (1-\theta) \|u_2\| \le \theta + (1-\theta) = 1$. Hence, $x \in \mathcal{N}$, which proves convexity. However, for the special case of $\|\cdot\|_p$, the triangular inequality is not satisfied unless $1 \le p$; this completes the proof. ■

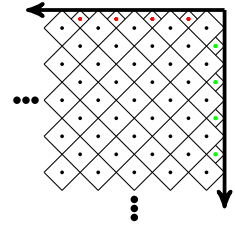

Figure 12: Partitioning the region $]-\infty, i] \times ]-\infty, j]$, with neighborhoods under $L^1$. The black, red, and green centroids belong to partitions $\mathcal{R}_r, r = 1, 2, 3$, respectively, in the proof.

**Lemma 6.** *The sample NCDF* (3) *converges pointwise to the population NCDF* (2a) *as $n \longrightarrow \infty$.*

**Proof.** This is trivial, but we provide it for the sake of completeness. It is known from the strong law of large numbers (S.L.L.N.) that $\frac{1}{n} \sum_i I_i$, $i = 1, \ldots, n$, for any $I \sim \text{Ber}(p)$, converges a.s. to $p$. Therefore, given a particular value of $v$, the L.H.S. of (3) converges a.s. to $\Pr[\mathcal{N}_{\|\cdot\|_p}(x_0, V_{\|\cdot\|_p}^{-1}(v))]$; and because the choice of $v$ is arbitrary, the whole $\overline{NCDF}_X$ curve converges pointwise to $NCDF_X$. ■

**Lemma 7.** *Both $NCDF_{\|\cdot\|_p, X=x}$ and $F_X$ are equivalent for the cases: $p = \infty, \forall d$, and $p = 1$, $d = 1, 2$.*

**Proof.** Under a continuous probability measure, all boundaries of open sets sum to a set of zero measure. *Necessity:* Recall that a point on the NCDF curve is a plot of $\Pr[\mathcal{N}\|\cdot\|_p(x, \epsilon)]$ vs. $\epsilon$. It is straightforward to show that any open set, including the open neighborhood $\mathcal{N}\|\cdot\|_p(x, \epsilon) \forall p$, is a Borel set that can be represented as a countable union of hyper cubes. Hence, its probability can be represented as an infinite summation in terms of the CDF function: $\Pr[\mathcal{N}\|\cdot\|_p(x, \epsilon)] = \sum_{i=1}^{\infty} F_X(x_i)$. *Sufficiency:* Recall that each point on the CDF is a plot of $F_X(x)$ vs. $x$. Given the NCDF family, we need to prove that the probabilities $\Pr[\mathcal{N}_{\|\cdot\|_p}(x, \epsilon)]$ $\forall x, \epsilon$ can be used to calculate $F_X(x) = \Pr[[-\infty, x]^d]$. We pursue this by showing that the region $[-\infty, x]^d$ can be represented as a countable partition of different neighborhoods; i.e., $[-\infty, x]^d = \bigcup_i \mathcal{N}_{\|\cdot\|_p}(x_i, \epsilon_i)$, where the neighborhoods are disjoint. First, for $p = \infty$, the neighborhoods are hyper cubes in $d$-dimensions. Hence, they naturally partition the region $[-\infty, x_0]^d$. Second, for $p = 1$, the case of $d = 1$ is trivial; and for $d = 2$, we partition $[-\infty, x]^2$ to three disjoint regions $\mathcal{R}_r, r = 1, 2, 3$; each region is a countable union of $\mathcal{N}\|\cdot\|_1(x, \epsilon)$ as follows (Figure 12): $\mathcal{R}_1 = \bigcup_{i,j=1}^{\infty} \mathcal{N}\|\cdot\|_1((i,j), 1)$, $i = n$, $j = 2m - (n \mod 2), \mathcal{R}_2 = \bigcup_{i,j=1}^{\infty} \mathcal{N}\|\cdot\|_1((i,j), i)$, $i = 2^{-n}$, $j = I_{n=1} + 2^{1-n}(2m-1), \mathcal{R}_3 = \bigcup_{i,j=1}^{\infty} \mathcal{N}\|\cdot\|_1((i,j), j)$, $j = 2^{-n}$, $i = I_{n=1} + 2^{1-n}(2m-1)$, where $n, m = 1, 2, \cdots$. ■

**Lemma 8.** *Consider the dataset $\mathbf{X} = \{x_i \mid i = 1, \ldots, n\}$ that exists in the $d$-dimensional unit cube $[0, 1]^d$, along with the two general observations $x_i, x_j \in \mathbf{X}$, and the distance $\epsilon = \|x_i - x_j\|_p$. Then, under the norm $\|\cdot\|_p$ $\forall p$, the following holds:*

*1) The maximum distance between $x_i$ and $x_j$ is $\epsilon_{max} = \max_{i,j} \|x_i - x_j\|_p = d^{1/p}$, which decreases monotonically with $p$, and occurs iff the two observations are at two opposite vertices of the cube; i.e., $x_i = (I_1, \ldots, I_d)$, $x_j = 1 - x_1$, $I_{d'} = 0, 1$, and $d' = 1, \ldots, d$.*

*2) The volume $V_{\|\cdot\|_p}(\epsilon)$ of the neighborhood $\mathcal{N}_{\|\cdot\|_p}(x_i, \epsilon)$ is maximized, iff $\epsilon = \epsilon_{max}$, and is given by*

$$V_{max} = (2d^{1/p})^d \frac{\Gamma(1 + 1/p)^d}{\Gamma(1 + d/p)}. \qquad (4)$$

*3) The ratio between the volumes of two neighborhoods or radii, $\epsilon$ and $\epsilon'$ is $(\epsilon/\epsilon')^d$; and if $\epsilon' = \epsilon_{max}$ (normalization by max. neighborhood), the ratio is $\left(\epsilon/d^{1/p}\right)^d$.*

**Proof.** (1) $\epsilon = \|x_i - x_j\|_p = \left(\sum_{d'} |(x_{id'} - x_{jd'})|^p\right)^{1/p}$, which takes its maximum when $|(x_{id'} - x_{jd'})|$ takes its maximum $\forall d'$. This happens when $x_{id'} = 1$ and $x_{jd'} = 0$ or vice versa. Hence, $\epsilon_{max} = d^{1/p}$, which is monotonically decreasing with $p$. (2) From [58], $V_{\|\cdot\|_p}(\epsilon) = (2\epsilon)^d \frac{\Gamma(1+1/p)^d}{\Gamma(1+d/p)}$, which is monotonic in $\epsilon$—from the properties of the Gama function $\Gamma$—hence possessing its maxima at $\epsilon_{max}$. (3) The proof is immediate from the ratio $V_{\|\cdot\|_p}(\epsilon)/V_{max}$. ∎

**Lemma 9.** *Consider the r.v. $X \in [0, 1]$ and the observations $X = \{x_i | i = 1, \cdots, n\}$ (to represent the intercepts of a $\beta$-level with the $n$ NCDF curves), and consider a constant point (intercept) $x \notin X$. The distance $\|x - X_k\|$, from the point $x$ to its closest neighbor $x_{(k)}$, converges to zero almost surely, as $n \to \infty$ unless $k/n \to r$, where $0 < r \leq 1$ is an arbitrary constant.*

**Proof.** The proof is similar to the proof of Lemma 5.1 in [59]. For any $0 < \epsilon < 1$, we have the equivalence $\|x - X_k\| > \epsilon \equiv \sum_i I_{X_i \in \mathcal{N}(x, \epsilon)} < k$. Because $\frac{1}{n} \sum_i I_{X_i \in \mathcal{N}(x, \epsilon)}$ converges a.s. to $\Pr(\mathcal{N}(x, \epsilon))$, then if $k/n \to 0$ this probability will be zero and $\Pr(\|x - X_k\| > \epsilon)$ converges to zero as well. Therefore, $k/n$ must converge to a constant $r$ (or as a special case $k = nr$) so that $\|x - X_k\|$ converges to a non-zero value. ∎

**Waleed A. Yousef**, Ph.D. (2007) in computer engineering, M.Sc. in mathematical statistics, George Washington University, U.S., and an M.Sc. in computer science, Helwan University. He worked as a research fellow at the U.S. FDA in machine learning and pattern recognition. Currently, he is a senior scientist at ISOT lab., ECE Dep., University of Victoria (UVic), Canada. He is the founder of MESC Labs (www.mesclabs.com).

**Issa Traore**, Ph.D. (1998) in Software Engineering, Institute Nationale Polytechnique, France. He is currently a professor and the coordinator of the ISOT Lab. (http://www.isot.ece.uvic.ca), ECE dep., UVic; and an associate editor of IEEE, TIFS. His research interests include biometrics technologies, computer intrusion detection, network forensics, software security, and software quality engineering. He is the co-founder of Plurilock Security Solutions Inc. (http://www.plurilock.com) and Email Veritas Security Technologies Ltd. (https://www.emailveritas.com).

**William Briguglio**, M.Sc. (2020), B.Sc. with a minor in Mathematics with distinction and honor degree, both in computer science, University of Windsor. Currently, he is a Ph.D. student and a research assistant at ISOT lab., ECE dep., UVic.